\documentclass[journal]{IEEEtran}  % Comment this line out if you need a4paper
\usepackage[utf8]{inputenc}
\usepackage{subcaption}
\usepackage{graphicx}
\usepackage{tcolorbox}
\usepackage{textcomp}
\usepackage{romannum}
\usepackage{enumerate}
\usepackage{booktabs}
\usepackage{multirow}
\usepackage{flushend}
\usepackage{amsthm}

\usepackage{amsmath}
\usepackage{amsfonts}
\usepackage[sharp]{easylist}
\usepackage[T1]{fontenc}
\usepackage[separate-uncertainty = true, multi-part-units=single]{siunitx}
\usepackage{bm}
\usepackage{tikz}
\usepackage{pifont}
\usepackage{pgfplots}
\usepackage{adjustbox}
\usepackage[capitalize]{cleveref}

\usepackage{cite}

\crefformat{equation}{(#1)}
\crefrangeformat{equation}{(#1) to (#2)}
\crefmultiformat{equation}{(#1)}{ and (#1)}{, (#1)}{ and (#1)}
\crefrangemultiformat{equation}{(#1) to (#2)}{}{}{}

\theoremstyle{definition}
\newtheorem{definition}{Definition}[section]

\usetikzlibrary{patterns, backgrounds, intersections}
\pgfplotsset{compat=newest}
\usepgfplotslibrary{groupplots}
\usepgfplotslibrary{polar}
\usepgfplotslibrary{smithchart}
\usepgfplotslibrary{statistics}
\usepgfplotslibrary{dateplot}
\usepgfplotslibrary{ternary}
\usetikzlibrary{arrows.meta}
\usetikzlibrary{backgrounds}
\usepgfplotslibrary{patchplots}
\usepgfplotslibrary{fillbetween}
\pgfplotsset{%
layers/standard/.define layer set={%
    background,axis background,axis grid,axis ticks,axis lines,axis tick labels,pre main,main,axis descriptions,axis foreground%
}{grid style= {/pgfplots/on layer=axis grid},%
    tick style= {/pgfplots/on layer=axis ticks},%
    axis line style= {/pgfplots/on layer=axis lines},%
    label style= {/pgfplots/on layer=axis descriptions},%
    legend style= {/pgfplots/on layer=axis descriptions},%
    title style= {/pgfplots/on layer=axis descriptions},%
    colorbar style= {/pgfplots/on layer=axis descriptions},%
    ticklabel style= {/pgfplots/on layer=axis tick labels},%
    axis background@ style={/pgfplots/on layer=axis background},%
    3d box foreground style={/pgfplots/on layer=axis foreground},%
    },
}

\newcommand{\mat}[1]{\bm{#1}}
\renewcommand\vec{\bm}

\usepackage{tikz}

\IEEEoverridecommandlockouts                              % This command is only needed if 
\title{\LARGE \bf
Reacting to Contact: Transparency and Collision Reflex in Actuation
}

\date{December 2022}

\author{Ankit Bhatia$^{1}$, Matthew T. Mason$^{1}$, and Aaron M. Johnson$^{2}$% <-this % stops a space
\thanks{* This material is based upon work supported by the National Science Foundation under Grant IIS-1813920. }% <-this % stops a space
\thanks{$^{1}$ Ankit Bhatia and Matthew T. Mason are with the Robotics Institute, Carnegie Mellon University
        {\tt\small {ankitb, mason}@cs.cmu.edu}}%
\thanks{$^{2}$Aaron M. Johnson is with the Department of Mechanical Engineering, Carnegie Mellon University,
        {\tt\small amj1@andrew.cmu.edu}}%
}

\begin{document}
\bstctlcite{IEEEexample:BSTcontrol}

\maketitle
\thispagestyle{empty}
\pagestyle{empty}

\begin{abstract}

In unstructured environments, robots run the risk of unexpected collisions. How well they react to these events is determined by how transparent they are to collisions. Transparency is affected by structural properties as well as sensing and control architectures. In this paper, we propose the collision reflex metric as a way to formally quantify transparency. It is defined as the total impulse transferred in collision, which determines the collision mitigation capabilities of a closed-loop robotic system taking into account structure, sensing, and control. We analyze the effect of motor scaling, stiffness, and configuration on the collision reflex of a system using an analytical model. Physical experiments using the move-until-touch behavior are conducted to compare the collision reflex of direct-drive and quasi-direct-drive actuators and robotic hands (Schunk WSG-50 and Dexterous DDHand.) For transparent systems, we see a counter-intuitive trend: the impulse may be lower at higher pre-impact velocities.

\end{abstract}

\section{Introduction}

Robots are moving out of the structured environment of a factory and into more complicated environments like warehouses and homes. In these spaces, we would like robots to do tasks like small parts assembly, cooperative manipulation, and putting away the dishes.
But these new applications in cluttered spaces require more delicate intentional interaction with objects as well as safe unexpected contact with the world. \cite{goertz1963Manipulators} explains the problem well: ``General-purpose manipulation consists essentially of a series of collisions with unwanted forces, the application of wanted forces and the application of desired motions. The collision forces should be low and any other unwanted forces should be small."
This work aims to better characterize design and control of robots for a lighter touch.  

There are many ways to mitigate the effect of a collision. 
In manufacturing applications, robots tend to approach a possible contact event at slower speeds than when operating in free space to minimize impulses, e.g.~\cite{joseph2020Online,lasota2014Safe}. 
When more precision is needed, touch-based localization can be used to approach with a guarded move, i.e.\ a ``move until touch'' behavior \cite{will1975experimental,mason1981Compliance}.

Higher velocity guarded moves can cause large impacts and damage the work piece or the robot itself. Another challenge emerges when considering collaborative robots and their ability to handle contacts with humans. These robots cannot always rely on exteroceptive sensors to anticipate the presence of a human in their motion path. 
Similarly, in robotic locomotion, when a walking robot impacts the leg on the ground, high impacts will cause mechanical noise while walking and lower efficiency.

Being able to manage these impacts through design and control advances can lead to better performing behaviors. \cite{haddadin2013Safe} looks at design and control strategies with the KUKA LWR to make industrial manipulators safe in human-robot collisions through reactive strategies. \cite{bhatia2019Direct,ostyn2022design, saloutos2022fast} demonstrate high-speed grasping behaviors with low-inertia, high bandwidth grippers. Low inertia fingers are also useful for tactile exploration, contact localization and shape recovery~\cite{saloutos2022towards,saloutos2022fast, wang2020Contact,lin2021Exploratory}.

\begin{figure}
  \centering
  \begin{subfigure}[t]{0.49\columnwidth}
    \includegraphics[width=\textwidth]{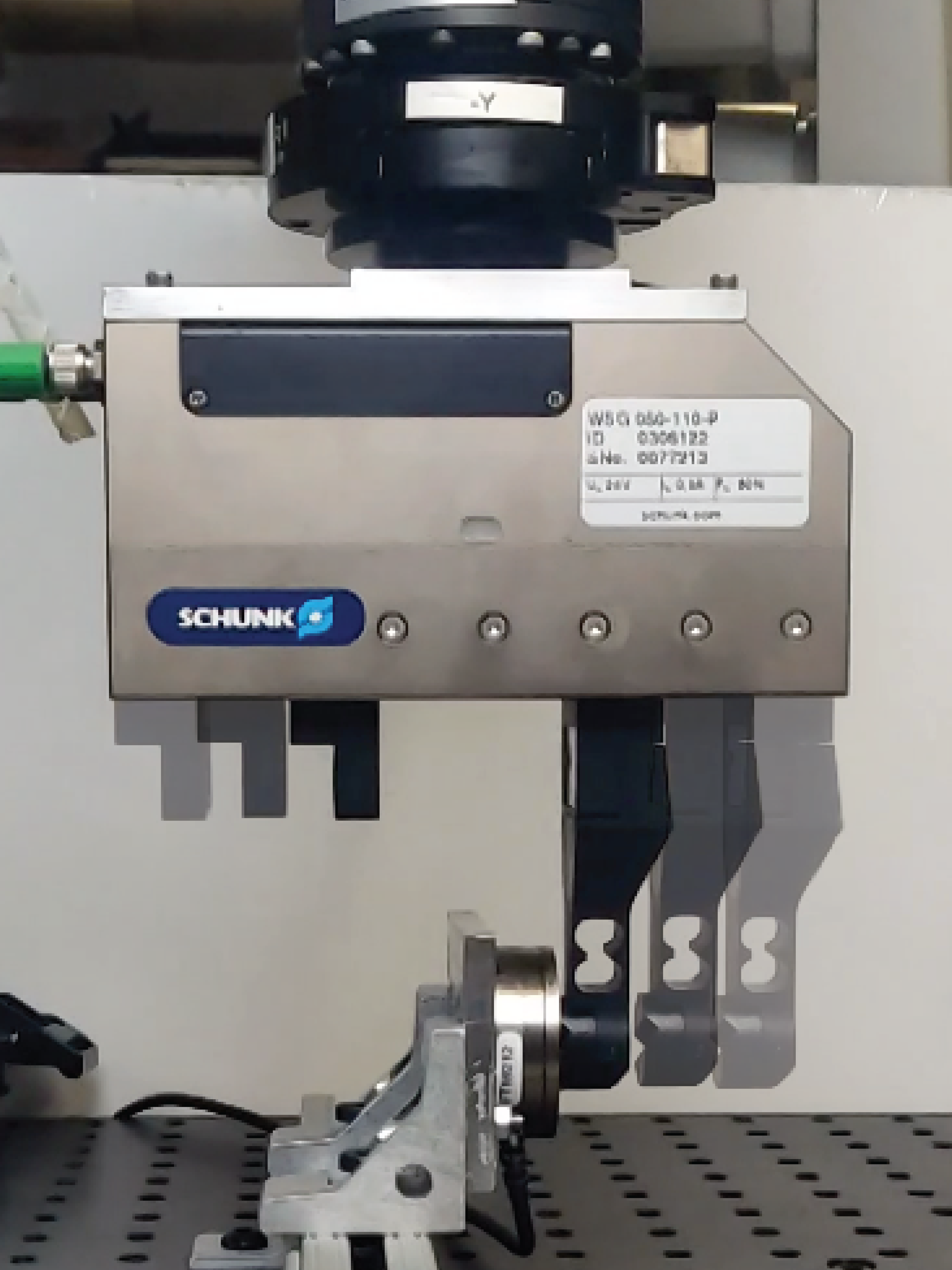}
    \caption{Schunk Gripper}
  \end{subfigure}
  \begin{subfigure}[t]{0.49\columnwidth}
    \includegraphics[width=\textwidth]{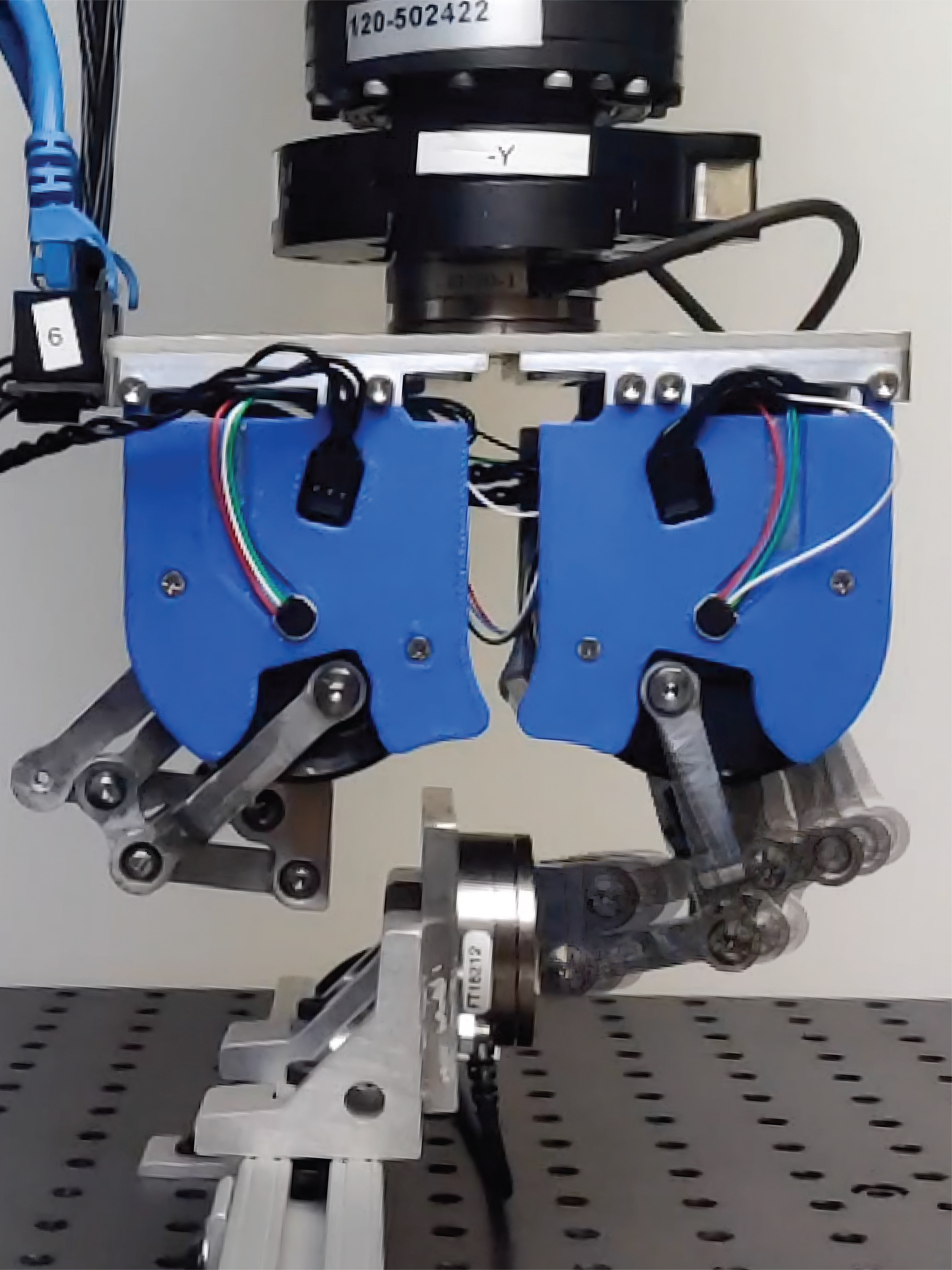}
    \caption{DDHand}
  \end{subfigure}
  \begin{subfigure}[t]{\columnwidth}
    \centering
    \includegraphics[width=\columnwidth]{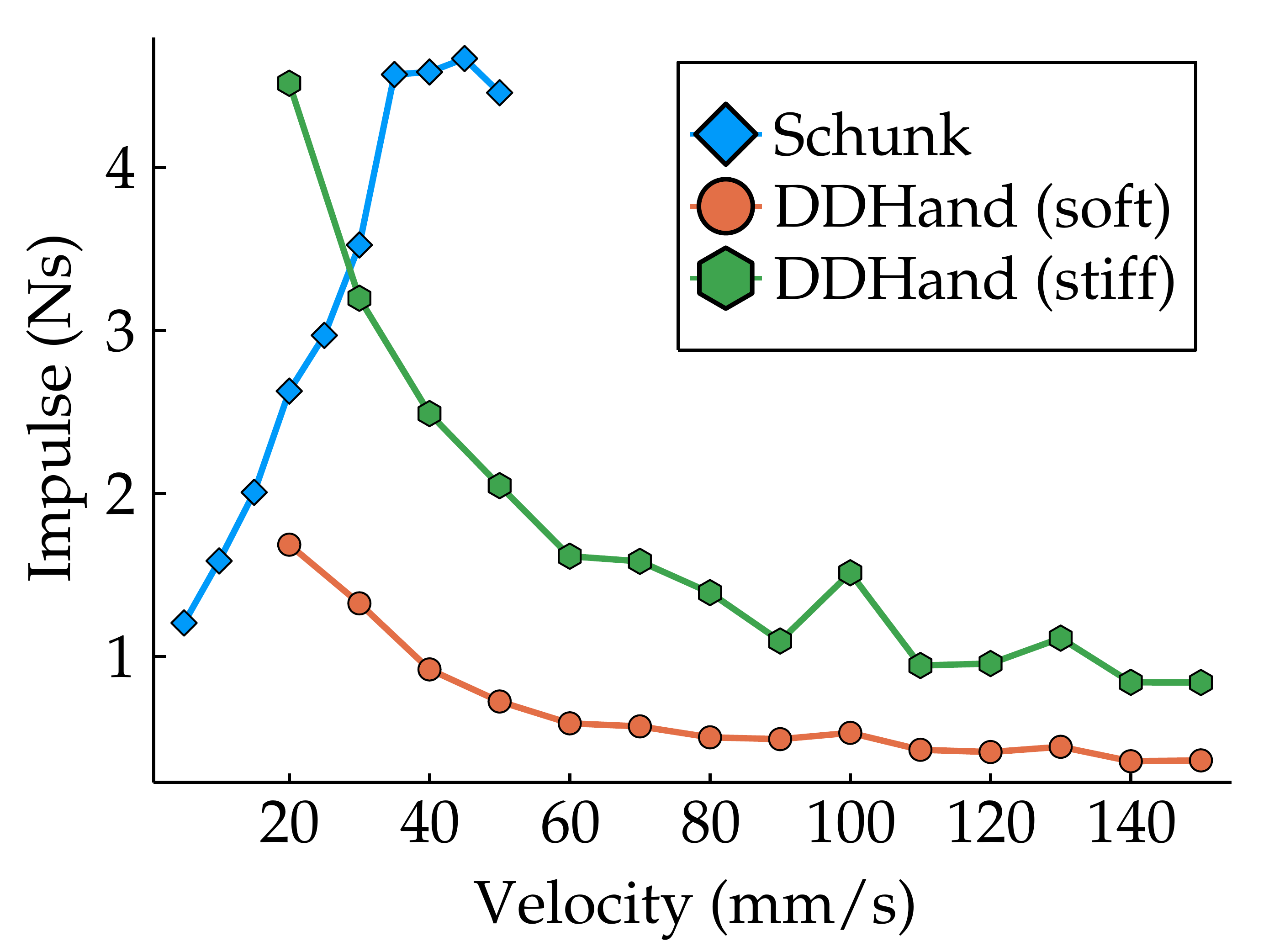}
    % \caption{Impule vs. Velocity for }
  \end{subfigure}
  \caption{The collision reflex experiment for the DDHand \cite{bhatia2019Direct} and Schunk end-effectors. (Top): The fingertips are impacted against a force sensor with a move-until-touch and react control loop. (Bottom): The total impulse for varying pre-impact velocities. }
  \label{fig:reflex_experiment}
\end{figure}

The transparency~\cite{kenneally2018Actuator,seok2012Actuator} of a system plays a role in how it manages the impulse during an unexpected collision event. Transparency captures the efficiency of transmission of force and motion between the task and the control agent. This idea of transparency is derived from teleoperation research where it is formally defined as the ratio of operator and task impedance~\cite{lawrence1993Stability}. 
No such formal definition exists for closed-loop robotic systems. Instead, we discuss the transparency properties of a system by looking at metrics that describe the inertial properties~\cite{seok2012Actuator} and closed-loop responses~\cite{kenneally2018Actuator} of the system.
Although these metrics have been useful for optimizing the design and control of robots to minimize impacts, no single metric captures the combined effect of inertia, sensing, and control. %TODO examples for useful design/control opt here.

We propose the collision reflex metric as a way to describe the transparency properties of the system. \cite{haddadin2013Safe} first discussed the collision reflex as a way to understand the response of a robot during an unexpected collision event. We extend this idea and analyze the total impulse transferred in such a collision event and discuss the impulsive contribution of the structure, sensing, and control. We analyze both a simple, one-dimensional example and extend this to higher dimensional systems, e.g. a two-link manipulator in the plane. 

% To understand the utility of the ideas presented in the paper, we explore a case study in manipulation. We introduce the Direct-drive hand and demonstrate its performance on a manipulation task -- the smack and snatch grasp. 
We then describe experiments to compare the total impulse transferred in collision of physical systems and assess the collision reflex metric for various actuator configurations and COTS grippers.

In summary, the contributions of the paper are: 
1) A metric which quantifies the collision reflex capability of a robotic system by measuring the total impulse imparted to a fixed rigid object;
2) An extension of this collision metric to cover cases that can fit the general manipulator equation and higher dimensional systems; 
3) Analysis of the effect of control, structure, and sensing on the total impulse transaction in a collision; and
4) Experimental results that compare the total impulse transferred in collision of physical systems and an assessment of the collision reflex metric for different actuation schemes.

\section{Related Work}
\label{sec:related}
Understanding a robot's reaction to contact requires a closer look at the robot's structure, sensing, and actuation paradigms. Robots come in a variety of shapes and sizes, with serial or parallel mechanisms; tactile, proprioceptive, or exteroceptive sensing modalities; hydraulic, electromagnetic, or pneumatic actuators that can be rotary or linear. 

% \begin{easylist}
%     # This paper will focus on electromagnetic actuators but ideas discussed in this paper are extendable to other forms of actuation
%     # When we think of a conventional electromagnetic actuator, a geared servomotor comes to mind. 
%     # More recently as we start exploring the space of collaborative robots and dynamic walking robots, we are expanding this definition to include ideas of proprioceptive sensing and compliance. 
% \end{easylist}

\subsection{General Models of Actuation} 
It is common to simplify the general servo-electric actuator as a mass-spring-mass system \cite{bhatia2019Direct,haddadin2013Safe,seok2012Actuator}. 
In \cite{bhatia2019Direct}, we proposed a first-principles model of an electric actuator to understand the bandwidth properties of different actuator schemes. The model consists of a motor with inertia $J_M$ connected through a gearbox with ratio N and inertia $J_G$, and a spring with stiffness $k$ connected to a load $J_L$. This model represents the most common schemes of electric actuation: series-elastic ($k$ is soft, $N$ is high); stiff, strain-gauge sensing ($k$ is stiff, $N$ is high); and direct-drive ($N$ is low, $k$ is stiff.) 

\subsection{Transparency} 
We adapt the term ``transparency'' to indicate the ease of bidirectional transmission of force and motion between some operator and the task.
Ideas of transparency originate in teleoperation where it means that the operator feels as if directly present in the task. Teleoperation researchers
\cite{hokayem2006Bilateral,hannaford1989design,lawrence1993Stability,salcudean2000Transparent} adopted models and analysis that transform the intuitive ``feeling present'' notion into conditions on the transmission of force and velocity between operator and task. 
These conditions require matching the impedance at the operator-controller interface to the impedance of the robot-task interface.

There are a few other metrics that are related closely to the concept of transparency. 
We cover these metrics in relation to the general manipulator equation with $n$ degrees-of-freedom and $m$ constraints,
\begin{equation}
  \vec{\tau} = \mat{M(q)}\vec{\ddot{q}} + \mat{C(q,\dot{q})\dot{q}} + \vec{N(q)} - \mat{J_c(q)}^T \vec{\lambda}.
\end{equation}
Here, the generalized forces are $\vec{\tau}\in\mathbb{R}^n$ and the generalized coordinates are $\vec{q}$, $\vec{\dot{q}}$, $\vec{\ddot{q}}\in \mathbb{R}^n$. The mass matrix, coriolis matrix, and gravity matrix are given by $\mat{M(q)}\in \mathbb{R}^{n\times n}$, $\mat{C(q,\dot{q})}\in \mathbb{R}^{n\times n}$, and $\vec{N(q)}\in \mathbb{R}^n$, respectively. Constraints are represented by the constraint Jacobian $\mat{J_c(q)}\in \mathbb{R}^{m\times n}$ and constraint forces are represented by $\vec{\lambda}\in \mathbb{R}^m$.

\subsubsection{Generalzied Inertia Ellipsoids}
The Generalized Inertia Ellipsoid~\cite{asada1983Control} ($\vec{u}^T\mat{\Lambda_0}\vec{u}$ for unit direction $\vec{u}$) expresses the resistance to changing the velocity of the end-effector in various direction for a human operator who holds the end-effector and applies a force with a fixed magnitude,
  \begin{equation}
    \mat{\Lambda_0} = \mat{J_c(q)}^{-T}\mat{M(q)} \mat{J_c(q)}^{-1}.
  \end{equation}
The Ellipsoid of Gyration~\cite{hogan1984Impedance}, $\vec{u}^T\mat{\Lambda_0}^{-1}\vec{u}$ can be substituted for this metric when the Jacobian $\mat{J_c(q)}$ is not invertible. 

\subsubsection{Dynamic Manipulability Ellipsoid}
The Dynamic Manipulability Ellipsoid~\cite{yoshikawa1985Manipulability, walker1994Impact}, $\vec{u}^T\mat{\Lambda_d}\vec{u}$ expresses the ease of changing the velocity of the end effector via the set of actuators that drive the manipulator joints by applying joint torques with a fixed magnitude,
  \begin{equation}
    \mat{\Lambda_d} = \mat{J_c(q)} (\mat{M(q)}^T\mat{M(q)})^{-1} \mat{J_c(q)}^T.
  \end{equation}

\subsubsection{Effective Mass Ellipsoid}
The Effective Mass Ellipsoid~\cite{khatib1995Inertial}, $m_e$ computes the effective mass along a direction $\vec{u}$ in a collision at the end-effector. 
\begin{equation}
  m_e = (\vec{u}^T\mat{\Lambda_0}^{-1}\vec{u})^{-1}.
  \label{eq:effective_mass}
\end{equation}

\subsubsection{Impact Mitigation Factor}
\label{sec:imf}
\cite{wensing2017Proprioceptive} propose the impact mitigation factor as a metric for the contribution of reflected actuator inertias to impacts applied to the system. The impact mitigation factor is based on the difference in impulses between the system ($\rho$) and when its joints are locked ($\rho_l$,) that is, infinite reflected inertias. 
The generalized inertia matrix for the free and locked systems is given by $\mat{\Lambda}$ and $\mat{\Lambda_l}$ respectively where, 
\begin{equation}
  \mat{\Lambda} = \mat{J_c(q)}^{-T} \mat{M(q)} \mat{J_c(q)}^{-1},
\end{equation}
\begin{equation}
  \mat{\Lambda_l} = \mat{J_c(q)}^{-T} \mat{M_{bb}} \mat{J_c(q)}^{-1},
\end{equation}.

Here $\mat{M_{bb}}$ is the inertia of the COM of the locked system. $\mat{M}$ can be partitioned into block matrices to separate out the locked dynamics from the internal joint dynamics as 
\begin{equation}
  \mat{M} = \begin{bmatrix}\mat{M_{bb}} & \mat{M_{bj}}\\
\mat{M_{jb}} & \mat{M_{jj}}\end{bmatrix}.
\end{equation}

Knowing $\rho = \mat{\Lambda} \vec{v}$ and solving for equal impact velocities in both systems, the impulse in the free system can be written in terms of the impulse in the locked system as 
\begin{equation}
  \rho = \mat{\Lambda} \mat{\Lambda_l}^{-1} \rho_l.
\end{equation}
Taking the difference of the impulses, 
\begin{equation}
  \rho_l - \rho = (\mat{I} - \mat{\Lambda}\mat{\Lambda_l}^{-1}) \rho_l.
\end{equation}
which gives us
$(\mat{I} - \mat{\Lambda} \mat{\Lambda_l}^{-1}$) as the impact mitigation matrix and its determinant defines the impact mitigation factor. An IMF of 1, corresponds to a system with perfect inertial backdrivability and tends towards the locked system when IMF approaches zero. 

\subsubsection{Feel-cage task}
\cite{kenneally2018Actuator} test the transparency of actuators with varying transmissions using a task designed to measure how the actuator feels the environment. The task involves contact detection and caging of balls with varying mass while fixing the impact velocity. Actuators with highly geared transmissions impart more energy to the balls resulting in failed caging. 

Although the feel-cage task does capture the effect of structure, sensing, and control on the transparency of a system, it has one shortcoming: the metric is binary -- either the ball is caged by the actuator or it isn't. To extend this to a more quantifiable metric, we look at the total impulse transaction in a collision as the collision reflex of a system. 
\section{Proposed Collision Reflex Metric}
\label{sec:collision_reflex_model}
\begin{figure}
  \begin{center}
    \usetikzlibrary{decorations.pathmorphing,patterns, positioning}
\begin{tikzpicture}[black!75]
	% Supporting structure
	\fill [pattern = north west lines] (-1.,0) rectangle ++(2,.2);
	\draw[thick] (-1.,0) -- ++(2,0);
	\node[draw,fill=yellow!60,minimum width=1cm,minimum height=0.5cm,anchor=north,label=east:$m_f$](M1) at (0,-1){};
	\draw[->, thick] (M1.north west) -- ++(0,0.3) node[above] {$v_0$};
	% Spring + Arrows
	\draw[] (M1.south) -- ++(0,-0.25) node[inner sep=0pt] (spring_start){};
	\draw[decoration={aspect=0.3, segment length=1.5mm, amplitude=2mm,coil},decorate] (spring_start) -- ++(0,-1.5) node[midway,right=0.25cm,black]{$k$} node[inner sep=0pt] (spring_end){}; 
	\draw[] (spring_end) -- ++(0,-0.3)node[coordinate](m21){} node[draw,fill=yellow!60,minimum width=1cm,minimum height=0.5cm,anchor=north,label=east:$m_r$](M2){};
	\draw[->, thick] (M2.north west) -- ++(0,0.3) node[above] {$v_0$};
	\node at (0,-4.5) {\Romannum{1}};
	\begin{scope}[xshift=2cm]
	% Supporting structure
		\fill [pattern = north west lines] (-1.,0) rectangle ++(2,.2);
		\draw[thick] (-1.,0) -- ++(2,0);
		\node[draw,fill=yellow!60,minimum width=1cm,minimum height=0.5cm,anchor=north](M1) at (0,0){};
		% Spring + Arrows
		\draw[] (M1.south) -- ++(0,-0.25) node[inner sep=0pt] (spring_start){};
		\draw[decoration={aspect=0.3, segment length=1.5mm, amplitude=2mm,coil},decorate] (spring_start) -- ++(0,-1.5) node[midway,right=0.25cm,black]{} node[inner sep=0pt] (spring_end){}; 
		\draw[] (spring_end) -- ++(0,-0.3)node[coordinate](m21){} node[draw,fill=yellow!60,minimum width=1cm,minimum height=0.5cm,anchor=north](M2){};
		\draw[->, thick] (M2.north west) -- ++(0,0.3) node[above] {$v_0$};
		\node at (0,-4.5) {\Romannum{2}};
	\end{scope}
	\begin{scope}[xshift=4cm]
		% Supporting structure
		\fill [pattern = north west lines] (-1.,0) rectangle ++(2,.2);
		\draw[thick] (-1.,0) -- ++(2,0);
		\node[draw,fill=yellow!60,minimum width=1cm,minimum height=0.5cm,anchor=north](M1) at (0,0){};
		% Spring + Arrows
		\draw[] (M1.south) -- ++(0,-0.25) node[inner sep=0pt] (spring_start){};
		\draw[decoration={aspect=0.3, segment length=0.5mm, amplitude=2mm,coil},decorate] (spring_start) -- ++(0,-.75) node[midway,right=0.25cm,black]{} node[inner sep=0pt] (spring_end){}; 
		\draw[] (spring_end) -- ++(0,-0.3)node[coordinate](m21){} node[draw,fill=yellow!60,minimum width=1cm,minimum height=0.5cm,anchor=north](M2){};
		\draw[->, thick] (M2.south) -- ++(0,-0.3) node[below] {$F$};
		\node at (0,-4.5) {\Romannum{3}};
	\end{scope}
	\begin{scope}[xshift=6cm]
		% Supporting structure
		\fill [pattern = north west lines] (-1.,0) rectangle ++(2,.2);
		\draw[thick] (-1.,0) -- ++(2,0);
		\node[draw,fill=yellow!60,minimum width=1cm,minimum height=0.5cm,anchor=north](M1) at (0,0){};
		% Spring + Arrows
		\draw[] (M1.south) -- ++(0,-0.25) node[inner sep=0pt] (spring_start){};
		\draw[decoration={aspect=0.3, segment length=1.5mm, amplitude=2mm,coil},decorate] (spring_start) -- ++(0,-1.5) node[midway,right=0.25cm,black]{} node[inner sep=0pt] (spring_end){}; 
		\draw[] (spring_end) -- ++(0,-0.3)node[coordinate](m21){} node[draw,fill=yellow!60,minimum width=1cm,minimum height=0.5cm,anchor=north,label=east:$m_r$](M2){};
		\draw[->, thick] (M2.south) -- ++(0,-0.3) node[below] {$F$};
		\draw[->, thick] (M2.south west) -- ++(0,-0.3) node[below] {$v_0$};
		\node at (0,-4.5) {\Romannum{4}};
	\end{scope}
\end{tikzpicture}
  \end{center}
  \caption{Robot collision with a rigid constraint: the three phases of impact: plastic impact phase, (\Romannum{1}\textrightarrow\Romannum{2}), the sensing phase, (\Romannum{2}\textrightarrow\Romannum{3}) and the reaction phase, (\Romannum{3}\textrightarrow\Romannum{4}). }
  \label{fig:msm}
\end{figure}
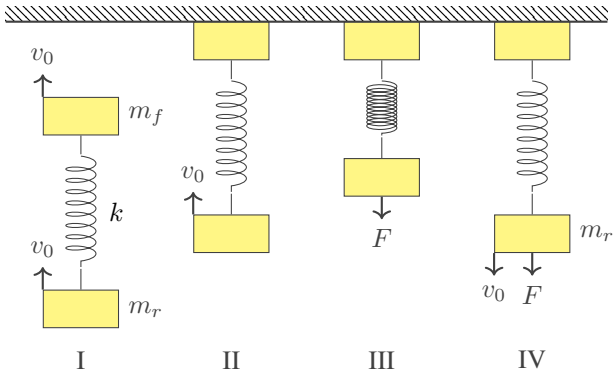

We define the collision reflex of a robot on its performance when it detects, decelerates and moves away from an unexpected collision. The more transparent the robot, the better it's performance in the collision reflex task.
The impulse in the collision is used as a metric to determine the quality of the collision reflex of the robot. The classical definition of impulse is used -- the integral of force over a time interval $t$ for which it acts.

\subsection{1D Model of Collision}
First, we analyze the impulse in a one-dimensional collision. For this collision model, we start with a model similar to the general actuator described in \cref{sec:related}. 
The collision model consists of a finger mass $m_f$ and a robot mass $m_r$ linked together by a spring of stiffness $k$. This spring lumps together the mechanical stiffness $k_m$ (from the structure and any physical springs present), and the software-defined springs $k_s$ in the controller. The difference between these two springs is that the rest length and stiffness of the software-defined spring is variable. The unsprung structure, transmission, and rotor inertias of the robot are lumped into the robot mass $m_r$ while all the sprung mass is lumped into $m_f$. The system approaches a constraint at a pre-impact velocity of $v_0$ and is capable of applying a force $F_a$ to the robot mass which causes an acceleration of $a = F_a/m_r$.

We model the robot's reflex to an unexpected collision over four events as shown in \cref{fig:msm}:
\begin{enumerate}
\renewcommand{\labelenumi}{\Roman{enumi}}
  \item The 1D finger-spring-robot system approaches a constraint with a pre-impact velocity of $v_0$. 
  \item At time $t_0 = 0$, $m_f$ loses all energy to the constraint in a plastic impact. As $m_r$ is still approaching the constraint with velocity $v_0$, it starts compressing the spring. We assume the robot is velocity controlled for this stage.
  \item When a collision event is detected at time $t_1$, the robot switches to a force controller resulting in an acceleration $a$ of the robot mass away from the constraint. 
  \item The collision event completes at time $t_2$ when the spring is not applying any more force on the constraint.
\end{enumerate}

\subsection{Phases of Impact}
We analyze the collision in 3 phases: plastic impact phase, $I_{\Romannum{1}\to\Romannum{2}}$, the sensing phase, $I_{\Romannum{2}\to\Romannum{3}}$ and the reaction phase, $I_{\Romannum{3}\to\Romannum{4}}$. The collision reflex impulse  is then the sum of the impulse contribution from each phase:
\begin{equation}
  I = I_{\Romannum{1}\to\Romannum{2}} + I_{\Romannum{2}\to\Romannum{3}} + I_{\Romannum{3}\to\Romannum{4}}.
\end{equation}

\begin{figure}
  \centering
  \includegraphics[width=\columnwidth]{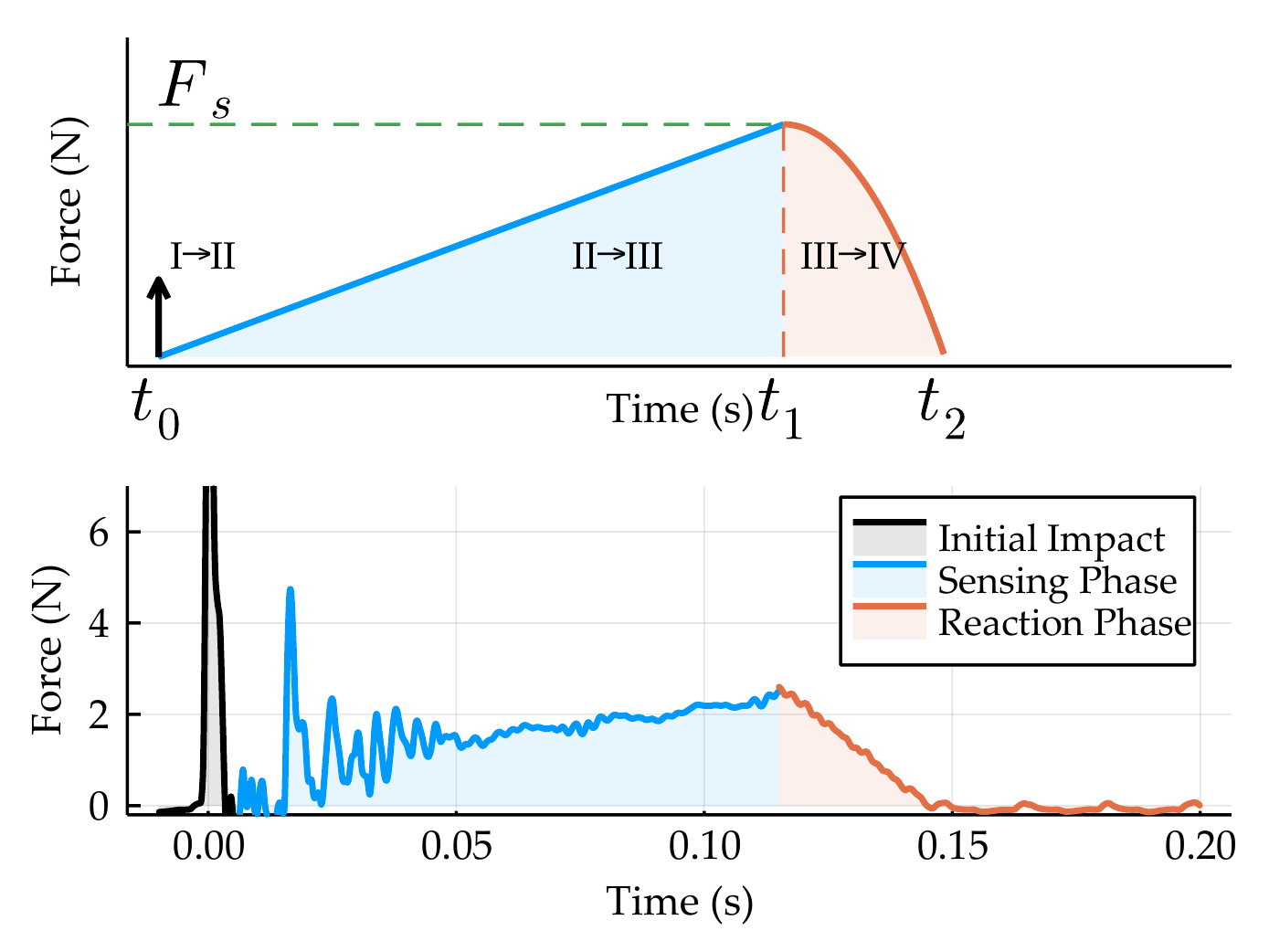}
  \caption{The three phases of impact are shown for the model as a force-time plot with significant features and time-points labeled (top.) Experimentally collected force-time profile of a collision for a U8 motor (bottom.) Details of the experiment can be found in \cref{sec:experiments}.}
  \label{fig:collision_reflex_plot}
\end{figure}

\subsubsection{Plastic Impact Phase} In this phase, the finger mass collides with the constraint and dissipates all of the momentum it carries. The impulse applied in this phase is thus the change in momentum of the finger as it comes to a instantaneous stop: 
\begin{equation}
  I_{\Romannum{1}\to\Romannum{2}} = m_f v_0.
  \label{eq:initial}
\end{equation}

\subsubsection{Sensing Phase} 
At the end of the plastic impact phase, system still has no information about the contact event and proceeds as usual. In this analysis we assume that the usual behavior of the system is to track a velocity reference for the robot mass. That is, until a collision is sensed, the robot mass is attempting to maintain a constant velocity $v$. 

This velocity controller presents as a ramp signal in the force-time signature of the collision up to the point of contact detection at time $t_1$. In this model the point of contact detection is represented by a threshold $F_s$. 

To compute the impulse $I_{\Romannum{2}\to\Romannum{3}}$ which is contributed during this stage, we first solve for the collision sensing time $t_1$: 
\begin{equation}
  t_1 = \frac{F_s}{k v_0}. \label{eq:t1def}
\end{equation}

The impulse in this phase is the area of the triangle shaded blue in \cref{fig:collision_reflex_plot}, 
\begin{equation}
  \begin{aligned}[b]
    I_{\Romannum{2}\to\Romannum{3}} &= \frac{1}{2} F_s t_1
        = \frac{F_s^2}{2k v_0} \\
  \end{aligned}
    \label{eq:sensing}
\end{equation}

\subsubsection{Reaction Phase} 

Once contact is detected, the robot is commanded an acceleration $a$ away from the constraint. If the robot mass $m_r$ is not at rest at the end of the sensing phase, part of the reaction phase would be used to decelerate the robot mass. In our analysis, we assume the contribution of this phase to be zero i.e. the reaction phase starts with the robot mass at zero velocity. The stiffness of this model has a discrete change to include only the mechanical stiffness of the robot and any software-defined stiffnesses or gains are instantaneously zeroed. 

Note that up until now there were two sources of stiffness: one from the mechanical elements (flex in the structure, strain gauges, fingertips/skin, etc) and one from software elements like control gains. The stiffness from the mechanical elements is fixed while the software stiffness is variable. We assume that elasticity from software with variable stiffness and rest length can be instantaneously zeroed. Thus, the only stiffness that needs to be overcome by the reaction phase is then the mechanical stiffness.

The acceleration command is assumed to start promptly after detection of collision at $t_1$ and ends at $t_2$, the time at which the spring has returned to its resting length. To compute the impulse $I_{\Romannum{3}\to\Romannum{4}}$ for this phase, we first need to compute $t_2$. 
This can be achieved by studying the motion of the robot mass starting at a displacement $F_s/k_m$ with zero velocity and ending at zero displacement. Using the laws of motion $s = ut + \frac{1}{2}at^2$ where $s$ is the displacement in the mechanical compliance, $u$ is the initial velocity (which is zero in this case,) $a$ the acceleration, and $t$ is the time of motion. Solving for the time of motion, we get
$t_2 = t_1 + \sqrt \frac{2F_s}{ a k_m}$.
The force during the reaction phase follows the profile defined by $F_2 = F_s - k_mat^2/2$
Integrating $F_2$ from $0$ to $t_2 - t_1$, we get the impulse during the reaction phase: 
\begin{equation}
    \begin{aligned}[b]
      I_{\Romannum{3}\to\Romannum{4}} &= F_s (t_2 - t_1) - \frac{1}{6} k_m a(t_2 - t_1)^3
%          &= {2 \over 3} F_s \sqrt{2 F_s \over a k_m}.
= \sqrt{\frac{8 F_s^3}{9ak_m}}.
    \end{aligned}
    \label{eq:reaction}
\end{equation}

Cases where the contribution of the deceleration of the robot mass plays a prominent role in the collision reflex are easy to account for. The initial velocity $u$ is no longer zero and is informed by the leftover momentum of the robot mass after the sensing phase. This would lead to changes in the expressions for $u$, $t_2$ and $I_2$ leading to added complexity in the collision reflex. As we will see in our experimental validation (~\cref{sec:experiments}), the reaction phase does not contribute prominently to the collision reflex. We can therefore safely assume that the robot mass is at rest at the end of the sensing phase.

\subsection{Total Impulse in a Collision}
After solving the contribution of each phase to the total impulse, we can add ~\cref{eq:initial}, \eqref{eq:sensing}, and \eqref{eq:reaction} to determine the total impulse $I$.
\begin{definition}[Collision Reflex Metric]
\begin{equation}
  I(m_f, F_s, k, k_m, v_0, a) = m_f v_0 + \frac{F_s^2}{2 k v_0} + 
  %{2\over 3} F_s \sqrt{2 F_s \over a k_m}.
  \sqrt{\frac{8 F_s^3}{9ak_m}}
  \label{eq:total_impulse}
\end{equation}
\end{definition}
The collision reflex metric depends on the finger mass $m_f$, sensing force $F_s$, stiffness $k$, pre-impact velocity $v_0$ and maximum allowable acceleration $a$.
  
The collision reflex assumes that the dynamics of the initial impact are much faster than the sensing bandwidth. If this is not true, and the sensor can trigger a collision detection event on the initial impulse, the sensing phase gets absorbed into the initial impulse. The robot mass at the end of the sensing phase can no more be assumed to be at rest. 

\subsection{Pre-impact Velocity for Minimum Impulse}

The analytical expression for the impulse in a collision for a generalized actuator shows dependence of velocity for the impact and sensing phases. The total impulse $I$ in \cref{eq:total_impulse} shows that the reaction phase impulse $I_{\Romannum{3}\to\Romannum{4}}$ does not depend on the pre-impact velocity of the system. Differentiating $I$ twice with respect to $v_0$, 
\begin{equation}
  \frac{\partial^2I}{\partial v_0^2} = \frac{F_s^2}{k v_0^3}.
\end{equation}
  
As $F_s^2$ and $k$ cannot be negative, ${\partial^2 I/\partial^2 v_0} > 0$ for all positive pre-impact velocities and the function $I$ is convex and has a single minima. The optimal velocity for minimal impulse (${\partial I/ \partial v_0} = 0$) is 
\begin{equation}
  v_0^* = \frac{F_s}{\sqrt{2k m_f}}.
  \label{eq:v_star}
\end{equation}

The minimum impulse at the optimal impact velocity $v^*$ is 
\begin{equation}
  I^* = \frac{F_s \sqrt{2}}{\omega_s} 
  %\left({\sqrt{2} + 1 \over \sqrt{2} }\right)
  + \frac{ F_s}{\omega_a} \sqrt\frac{8 F_s}{9 F_a},
\end{equation}
where $\omega_s = \sqrt{k/m_f}$, the open-loop sensing bandwidth (natural frequency) and $\omega_a = \sqrt{k/m_r}$ the open-loop actuation bandwidth of the system.

This analysis highlights relationships between the properties and behavior of the system during collision with a rigid body fixed in the world. As the stiffness of the system increases, the system behavior shifts towards an ideal rigid body collision, i.e. minimum impulse is at zero pre-impact velocity. 
We also see an inverse relationship between the open-loop bandwidth $\omega_a$ of the finger-mass and spring system and the impulse. As the open-loop sensing bandwidth goes up, the impulse reduces. The reaction phase is dependent on the open-loop actuation bandwidth of the system $\omega_a$. As the open-loop actuation bandwidth increases, the acceleration capability of the robot improves reducing the impulse applied during the reaction phase. 
% Thle choice of sensing threshold, $F_s$ is informed by the noise floor of contact sensing. 

The collision reflex allows for a quantitative measure of the transparency of the system which is only dependent on system parameters like pre-impact velocity, inertia, acceleration, and stiffness. Using this tool, we can directly compare the performance of two systems.

\section{Actuator Selection and the Collision Reflex }
\label{sec:motor_scaling}

With the ability to compare transparency of systems using the collision reflex metric, we can start to understand the tradeoffs between actuation schemes. Here we consider a space of actuation schemes spanned by two axes: stiffness, and reflected inertia. There are 4 quadrants of this space: high stiffness, high reflected-inertia (e.g.\ gearmotor + strain-gauge); low stiffness, high reflected-inertia (e.g.\ gearmotor + series-elastic actuation~\cite{pratt1995Series}); high stiffness, low reflected-inertia (e.g.\ direct-drive \cite{asada1983Control}); low stiffness, low reflected-inertia (e.g.\ compliant direct-drive~\cite{bauer2022Very,ebner1995Directdrive}). In this section, we look at actuation models to understand the effect on the collision impulse with changes in the reflected inertia and torque output of an actuator with one degree of freedom.

\subsection{Actuator Scaling}
\label{subsec:actuator_scaling_laws}

Torque and inertia of an actuator are closely related. Choosing a bigger motor for higher torque output will also increase its rotor inertia. A smaller motor on the other hand will produce insufficient torque for the application requiring a transmission. Reflected inertia of a gear motor scales the rotor inertia by a factor of $N^2$. 

% Let's look at an application which has a minimum torque requirement. How does the reflected inertia scale as we scale the motor? 

Various empirical and analytical models show that the scaling between rotor inertia $J_m$ and torque $\tau_m$ lies somewhere between linear and quadratic depending on assumptions. If isometric scaling is assumed~\cite{spletzer1999Scaling}, i.e. the length of the motor scales with the radius, $J_m \propto \tau_m$. Empirical analysis of motors~\cite{dermitzakis2011Scaling} show $J_m \propto \tau_m^{1.32}$ assuming inertia and mass follow isometric scaling. Quadraped design studies assume fixed radial thickness and motor length~\cite{seok2012Actuator,kenneally2016Design} to arrive at $J_m \propto \tau_m^{1.5}$, or assume fixed radial thickness and mass~\cite{seok2015Design} to arrive at $J_m \propto \tau_m^2$. 
Lastly, deriving the scaling laws based on electrical and thermal dynamics \cite{haddadin2012Rigid} results in $J_m \propto \tau_m^{1.6}$. 

\begin{table}
  \centering
  \renewcommand\arraystretch{1.3}
  \begin{adjustbox}{max width=\columnwidth}
    \begin{tabular}{lccc}
      \toprule
      Scaling Law& Mass & Reflected Inertia & Torque \\
      \midrule
      \textbf{Motors}&&&\\
      Isometric~\cite{spletzer1999Scaling} & $lr^2$ & $lr^4N^2$ & $lr^4$\\
      Empirical$^\dagger$~\cite{dermitzakis2011Scaling} & $lr^2$ & $lr^4N^2$ & $lr^{2.8}$\\
      Quadruped design~\cite{seok2012Actuator,kenneally2016Design} & $lr$ & $lr^3N^2$ & $lr^2$\\
      w/ electrical \& thermal~\cite{haddadin2012Rigid} & $lr^2$ & $lr^{4}N^2 $ & $lr^{2.5}$\\
      \midrule
      \textbf{Gearboxes~\cite{saerens2019Scaling}}&&&\\
      Parallel Shaft  & $lr^2$ & $lr^4N^2a^{-1}$ & $lr^2a^{-1}$\\
      Planetary       & $lr^2$ & $lr^4N^2a^{-1}$ &$lr^2a^{-1}$ \\
      Harmonic Drive  & $lr^2$ & $lr^4N^2$ &$r^3$ \\
      Cycloidal Drive & $lr^2$ & $lr^4N^2$ &$r^4l^{-1}$        \\
      Ball Screw      & $lr^2$ & $lr^4$ &$r^3$        \\
      \bottomrule
    \end{tabular}
  \end{adjustbox}
  \caption{Scaling Laws for mass, inertia and torque of motors and gearboxes. Here $l,\ r$ are the length and radius of the motor/gearbox, $N$ is the transmission ratio of the gearbox with $a$ stages. $^\dagger$ assumes inertia and mass follow isometric scaling.}
  \label{tab:scaling_laws}
\end{table}

When smaller motors are employed, a transmission is added to achieve a torque requirement. The contribution of reflected inertia by these transmissions may not be negligible, e.g.~\cite{saerens2020Scaling} shows the scaling laws for parallel shaft and planetary gear trains, harmonic and cycloidal drives and ball screws. We ignore the effect of this contribution to simplify the analysis and consider only the reflected rotor inertia. 

The scaling laws for motors and gearboxes are summarized in \cref{tab:scaling_laws}.

\begin{table}
  \centering
  \renewcommand\arraystretch{1.3}
  \begin{adjustbox}{max width=\columnwidth}
    \begin{tabular}{@{}lcc>{\bfseries}cc@{}}
      \toprule
      Parameter                       & Unit         & M1               & M2            & M3\\
      \midrule
      Stator OD ($r_m$)               &\SI{}{mm}     & \num{10.0}       & \num{60.0}    & \num{100.0} \\
      Stator ID                       &\SI{}{mm}     & \num{6.95}       & \num{41.7}    & \num{69.5}  \\
      Rotor OD                        &\SI{}{mm}     & \num{4.83}       & \num{40.6}    & \num{48.30} \\
      Rotor ID                        &\SI{}{mm}     & \num{3.36}       & \num{31.0}    & \num{33.57} \\
      Stack Height ($l$)              &\SI{}{mm}     & \num{12.5}       & \num{12.5}    & \num{12.5}  \\
      Rotor Inertia ($J_m$)           &\SI{}{kg.m^2} & \num[output-exponent-marker = $\mathrm{E}$]{1.71e-8}    & \num[output-exponent-marker = $\mathrm{E}$]{2.21e-5} & \num[output-exponent-marker = $\mathrm{E}$]{1.71E-4}\\
      Cont. Torque ($\tau_c$)         &\SI{}{N.m}    & \num[output-exponent-marker = $\mathrm{E}$]{5.94e-3}    & \num{0.524}   & \num{1.879e0} \\
      Peak Torque ($\tau_m$)          &\SI{}{N.m}    & \num[output-exponent-marker = $\mathrm{E}$]{1.47e-2}    & \num{1.3}     & \num{4.661}\\    
      Gear Ratio ($N$)                &-             & \num{88.18}      & \num{1}       & \num{1}\\
      Inertia ($N^2 J_m$)             &\SI{}{kg.m^2} & \num[output-exponent-marker = $\mathrm{E}$]{1.33e-4}              & \num[output-exponent-marker = $\mathrm{E}$]{2.21e-5} & \num[output-exponent-marker = $\mathrm{E}$]{1.71E-4}\\
      Peak Torque Out ($N \tau_p$)    &\SI{}{N.m}    & \num{1.3}        & \num{1.3}     & \num{4.661}\\
      \bottomrule
    \end{tabular}
  \end{adjustbox}
  \caption{Motor parameters of the Celera Motion Omni+ motor (OPN-060-013-A, M2 in the table) used for analyzing the effect of actuation selection on the collision reflex. The scaled parameters for motors with stator OD of \SI{10}{mm} (M1) and \SI{100}{mm} (M3) are also shown.}
  \label{tab:motor_params}
\end{table}

\subsection{Effect of Actuator Selection}

How does the selection of an actuation scheme affect the collision reflex of a system? To understand the relationship, we select a direct-drive motor and scale the motor radius keeping the length and the minimum torque output constant. A transmission is added to overcome the lower torque capability of smaller motors. 
As scaling laws from~\cite{haddadin2012Rigid} shows better correlation to empirical data~\cite{saerens2020Scaling} than the Quadruped laws~\cite{seok2012Actuator,kenneally2016Design}, they are applied to determine the inertia and torque output of the actuator. We assume that the effects of added inertia from the transmission are ignored and that all generated actuator configurations are feasible. The parameters of the selected motor (Celera Motion Omni+ series \SI{60}{mm} frameless motor OPN-060-013-A) and two scaled motors are shown in \cref{tab:motor_params}. 

\begin{figure}
  \centering
  % \begin{subfigure}[b]{0.45\textwidth}
  %   \centering
  %   \includegraphics[width=\columnwidth]{figures/stiff-rad-impulse.pdf}
  %   % \caption{}
  %   % \label{fig:stiff-rad-impulse}
  % \end{subfigure}
  \begin{subfigure}[b]{0.45\textwidth}
    \centering
    \includegraphics[width=\columnwidth]{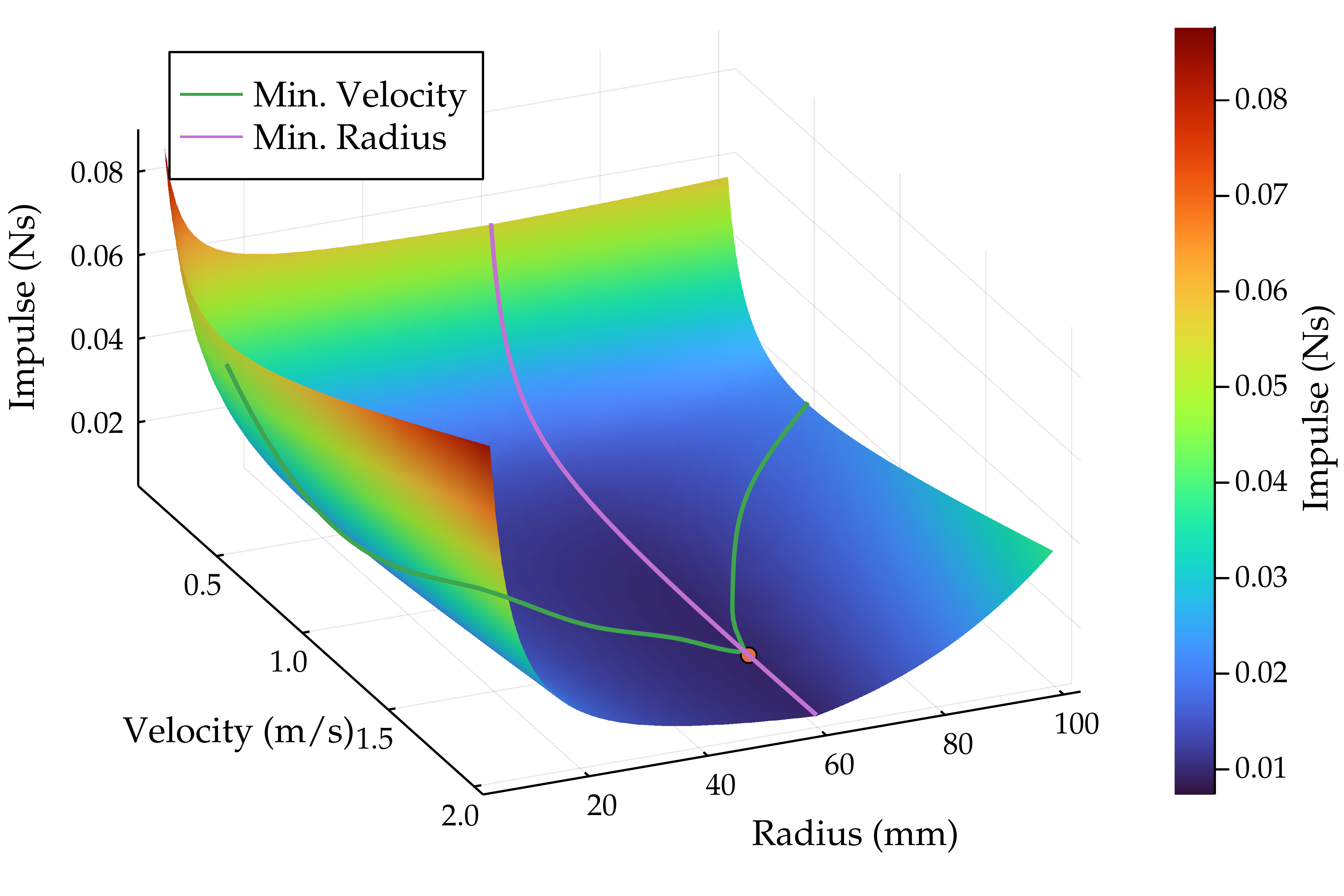}
    % \caption{}
    % \label{fig:vel-rad-impulse}
  \end{subfigure}
   \caption{Variation of the collision impulse as velocity and radius are varied with \SI{1000}{N/m} mechanical stiffness (right). The software stiffness $k_s$ was set at \SI{100}{N/m} with a collision sensing threshold $F_s$ of \SI{3}{N}. The orange dot on the plot shows the optimal parameters for the minimum impulse.}
   \label{fig:motor_scale_modeling}
\end{figure}

\cref{fig:motor_scale_modeling} shows the variation in collision reflex with changes in velocity and radius of the actuator.
Three operating ranges of motor scaling are seen. For small radius motors with high gear ratios, the impulse increases dramatically as the reflected inertia is scaled by the square of transmission ratio $N$. On the other end of the scale, large radius motors scale the impulse as the reflected inertia scales with the scaling law $J_m \propto r^{2.5}$. In the middle a region with low impulse is found which represents the quasi-direct-drive or direct-drive regime (low or unity transmission ratio) designs. Note that without the inclusion of reflected inertia added by the gearbox, optimal motor scaling lies somewhere in this range.

From \cref{eq:total_impulse}, the dependence of the impulse on velocity follows $I(v) = av + b/v + c$ where a, b and c are contributions of the three phases at unit pre-impact velocity, respectively. Lower velocities see the contribution of phase two dwarfs the other phases due to the slow ramp up to $F_s$. At higher velocities, phase one dominates when the inertia of the system is high, i.e. when the gear ratio is high or when larger than required motors are selected. These trends can be seen in \cref{fig:motor_scale_modeling}, with the optimal velocity lying somewhere in the middle.

\begin{figure}
  \centering
    \includegraphics[width=0.8\columnwidth]{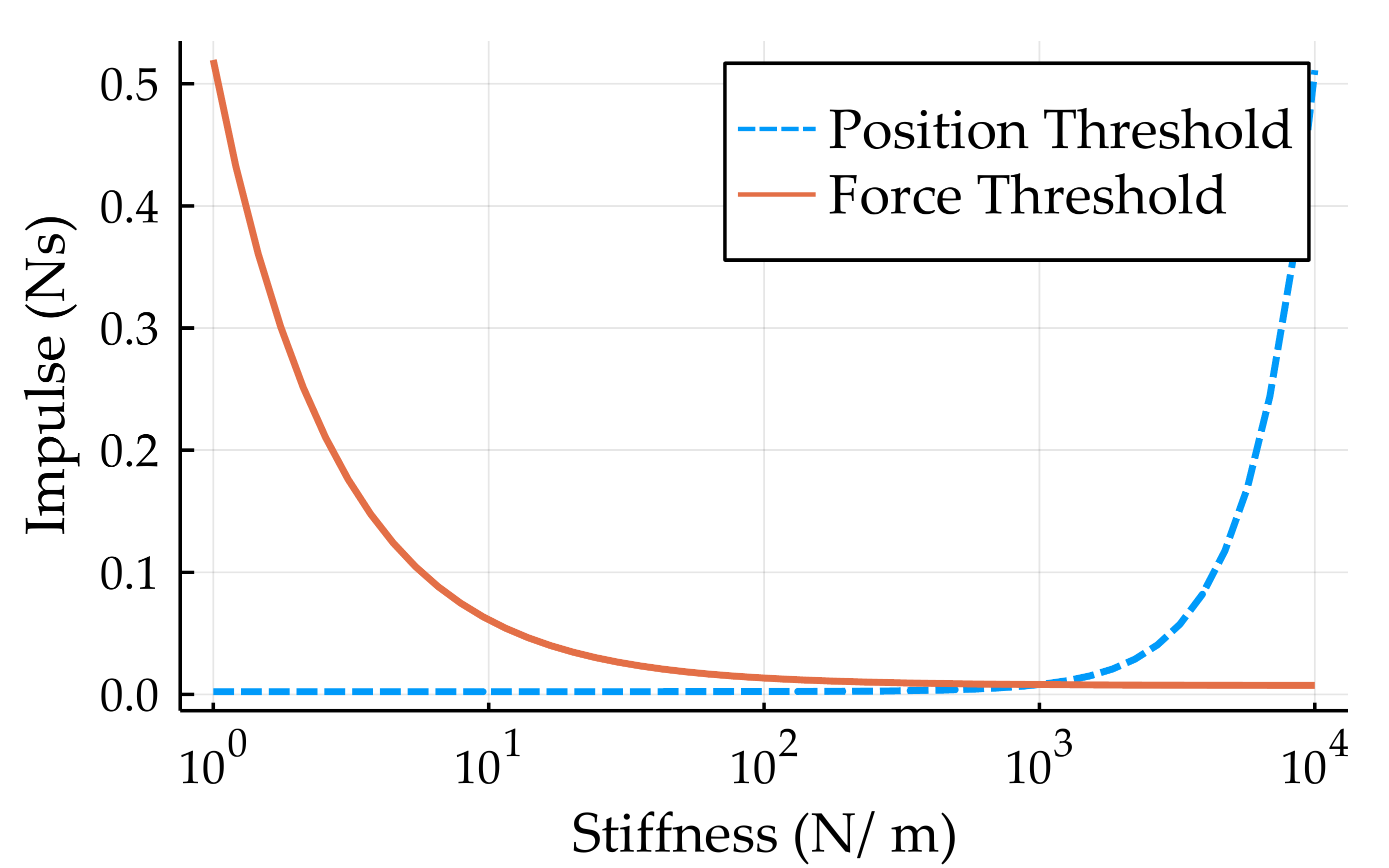}
    \caption{Effect of change in stiffness on the collision reflex impulse for two sensing schemes: position thresholding and force thresholding.}
    \label{fig:stiff-sensing-impulse}
\end{figure}

\cref{fig:stiff-sensing-impulse} shows the variation in collision reflex with changes in mechanical stiffness and sensing strategy. 
As the mechanical stiffness increases, the contribution of phases two and three decrease, \cref{eq:sensing} and \cref{eq:reaction}. At higher stiffness, the impulse reduces assuming the sensing force threshold $F_s$ remains constant. If the collision detection is implemented as a threshold on error in position control instead, the sensing threshold is proportional to the stiffness. This will cause the impulse to be higher as stiffness increases (as seen in \cref{sec:experiments}.)

\section{Collision Reflex in Higher Dimensions}

The collision reflex metric is also useful in design and planning tasks for higher dimensional systems. 

% \subsection{Model of Robot Collisions}
To this end, we extend the general manipulator described in \cref{sec:related}.
If we apply the actuator model discussed in \cref{sec:related} at each joint of the manipulator, we can decouple the actuator ($j$) dynamics from the structure ($b$). The new manipulator dynamics have a mass matrix as

% \begin{equation}
%   \mat{\tilde{M(q)}}\begin{bmatrix}
%     \vec{\ddot{q}} \\ \vec{\ddot{q}_a}
%   \end{bmatrix}+ 
%   \mat{\tilde{C}(q,\dot{q})}
%   \begin{bmatrix}
%     \vec{\dot{q}} \\ \vec{\dot{q}_a}
%   \end{bmatrix} + \vec{\tilde{N}(q)} + 
%   \mat{\tilde{K}}
%   \begin{bmatrix}
%     \vec{q}\\\vec{q_a}
%   \end{bmatrix} = 
%   \begin{bmatrix}
%     \vec{0}\\ \vec{\tau_a}
%   \end{bmatrix}
% \end{equation}

\begin{equation}
  \mat{\tilde{M}(q)} = 
  \begin{bmatrix}
    \mat{M_{bb}} & \mat{M_{jb}} \\ 
    \mat{M_{bj}} & \mat{M_{jj}}
  \end{bmatrix}.
\end{equation}
% \[
%   \mat{\tilde{C}(q,\dot{q})} = 
%   \left[\begin{smallmatrix}
%     \mat{C(q_0,\dot{q}_0)} & \mat{0}\\ \mat{0}&\mat{0}
%   \end{smallmatrix}\right],
% \] 
% \[
%   \vec{\tilde{N(q)}} = 
%   \left[\begin{smallmatrix}
%     \vec{N(q_0)} \\ \mat{0} 
%   \end{smallmatrix}\right],
% \] 
% \[
% \]
  
The dynamics of the structure and the actuators are coupled by a stiffness matrix 
\begin{equation}
  \mat{\tilde{K}} = 
  \begin{bmatrix}
    \mat{K} & -\mat{K} \\ 
    -\mat{K} & \mat{K}
  \end{bmatrix}, 
\end{equation}
where $\mat{K} = diag(k_1 \dots k_n)^T$ is a diagonal stiffness matrix with its diagonal elements describing the stiffness at each joint.

\begin{figure}
  \centering
  \includegraphics{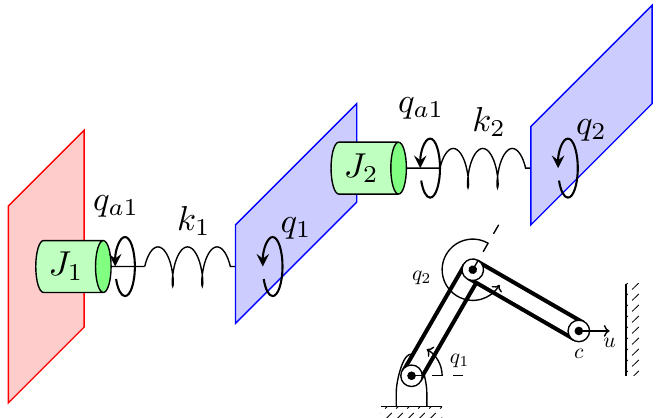}
  \caption{Schematic of a two-link manipulator colliding with a constraint. The quantities show the collision in the task space.}
  \label{fig:nlink_schematic}
\end{figure}

Let us look at the impulse transferred in a collision at the contact frame $c$ along a direction $u$, e.g.~as in \cref{fig:nlink_schematic}. We can apply the one dimensional collision reflex metric, \cref{sec:collision_reflex_model}, in this direction by using the directional effective mass, force, stiffness, and velocity in task space coordinates.
The initial impulse $I_{\Romannum{2}\to\Romannum{3}}$ is caused by the inertia $\mat{M_{bb}}$ which describes the structure of the robot. The impulse during phases two ($I_{\Romannum{3}\to\Romannum{4}}$) and three ($I_3$) are caused by the spring stiffness, $\mat{K}$ and the inertia of the actuators, $\mat{M_{jj}}$.

We can transform the inertias $\mat{M_{bb}}$ and $\mat{M_{jj}}$ 
and stiffness $\mat{K}$ defined in the generalized coordinate to task space coordinates through the Jacobian $\mat{J_c(q)}$. We can then map to the equivalent one-dimensional representation by projecting on the collision normal $\vec{u}$.
After applying \cref{eq:effective_mass}, the effective finger mass $m_f$, robot mass $m_r$, and stiffness $k$ at the contact point $\vec{c}$ in a collision direction $\vec{u}$ are
\begin{eqnarray}
  m_f &=& (\vec{u}^T (\mat{\Lambda})^{-1} \vec{u})^{-1},
  \label{eq:eff_mf}\\
  m_r &=& (\vec{u}^T (\mat{\Lambda_a})^{-1} \vec{u})^{-1}, \\
  k &=& \vec{u}^T \mat{K_t}  \vec{u}.
  \label{eq:eff_k}
\end{eqnarray}
where, $\mat{\Lambda} = \mat{J_c}^{-T} \mat{M_{bb}} \mat{J_c}^{-1}$ and $\mat{\Lambda_a} = \mat{J_c}^{-T} \mat{M_{jj}} \mat{J_c}^{-1}$ 
The task space stiffness matrix $\mat{K_t}$ can be computed through another Jacobian map as,
\begin{eqnarray}
  \vec{\tau}  &=& \mat{K} \Delta \vec{q},\\
  \mat{J_c}^T\vec{f}  &=& \mat{K} \mat{J_c}^{-1} \Delta \vec{x},\\
  \vec{f}     &=& (\mat{J_c}^{-T} \mat{K} \mat{J_c}^{-1}) \Delta \vec{x},\\
  \mat{K_t}   &=& \mat{J_c}^{-T} \mat{K} \mat{J_c}^{-1}. 
\end{eqnarray}

The task space acceleration capability of the robot $a$ is given by the Jacobian transpose map. 
\begin{equation}
  a = \frac{1}{m_r} \vec{u}^T \mat{J_c}^{-T} \mathrm{sat}(\mat{J_c}^T \vec{u}).
  \label{eq:eff_a}
\end{equation}
where the $\mathrm{sat}(\cdot)$ function saturates the motor torque to its maximum allowable limit. 

Having identified the one-dimensional collision parameters, $m_f$, $k$, and $a$ (Eqs.~\ref{eq:eff_mf}, \ref{eq:eff_k}, \ref{eq:eff_a} respectively), impulse for any collision frame $c$, pre-impact velocity $v_0$, force sensing threshold $F_s$, and impact direction $\vec{u}$ can be computed.

% \subsection{Example: 2 Link manipulator}

As an example, we apply the n-dimensional collision reflex extension to a 2 degree of freedom (DOF) manipulator. The manipulator shown in \cref{fig:nlink_schematic} is modeled with the Celera Motion motor discussed in \cref{sec:motor_scaling} to understand the effect of motor scaling in higher degree of freedom robots. 

\begin{figure*}
  \centering
  % \dummyfig{Two Link Manipulator }
  \includegraphics[width=\textwidth]{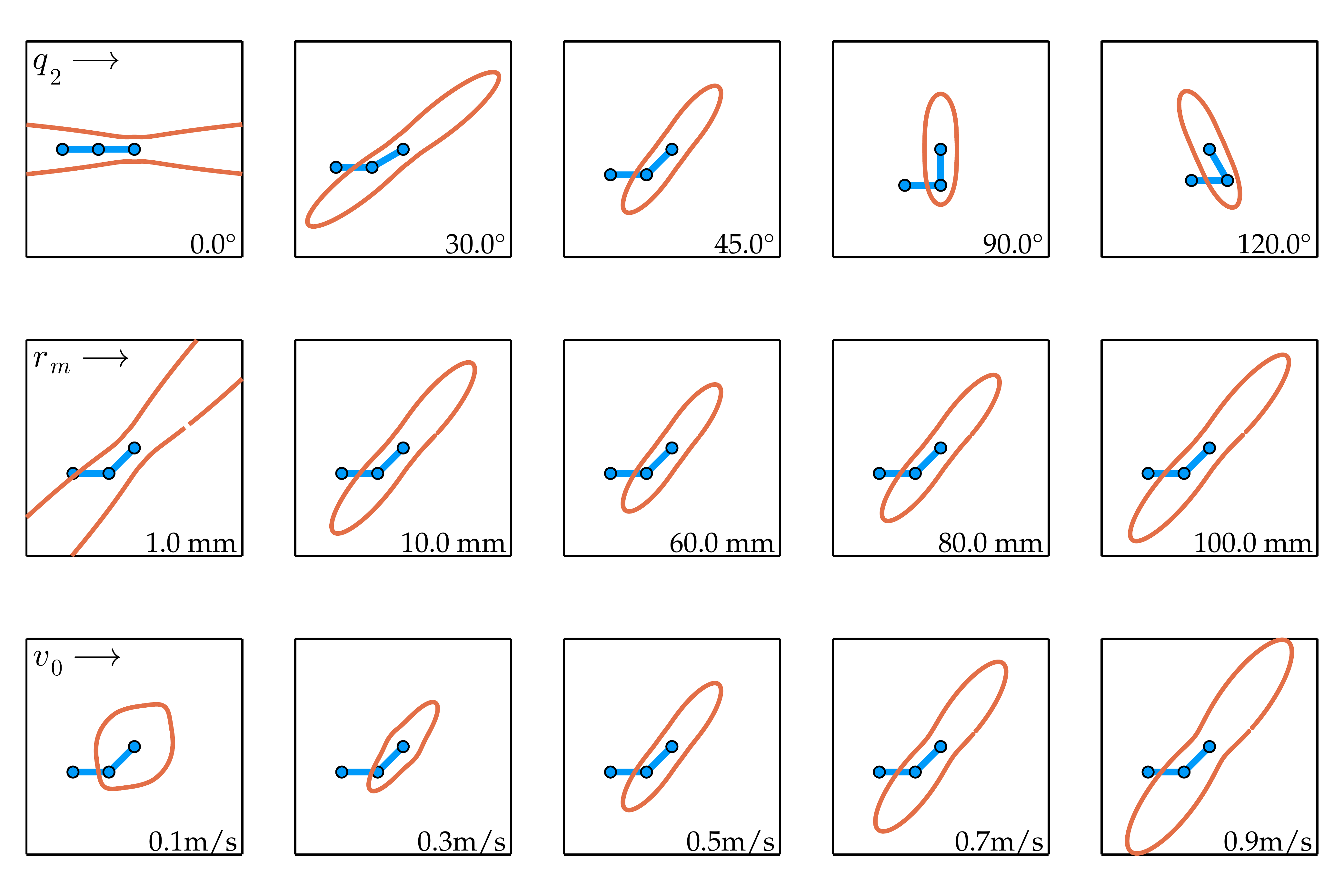}
  \caption{The collision reflex surface for a two link manipulator. (Top row) varying configuration; (middle row) varying motor scale; (bottom row) varying velocity. The robot is analyzed at $q_2 = 45^\circ$, $r_m = $ \SI{60.0}{mm}, and $v_0 = $ \SI{0.5}{m/s} unless otherwise indicated.}
  % in varying geometric configurations with the collision reflex surface overlaid for varying pre-impact velocities along the horizontal, varying configurations along the vertical and varying motor scaling in the plots.}
  \label{fig:two_link}
\end{figure*}

By sweeping vector $\vec{u}$ a full rotation about the end-effector, a collision reflex surface for a given configuration  of the manipulator is generated. \cref{fig:two_link} shows the collision reflex surface for a 2 DOF manipulator at various configurations $(q_1, q_2)^T$, motor scales $r_m$, and pre-impact velocities $v_0$.

As $q_2$ approaches 0, the collision reflex surface elongates to show the infinite stiffness along the singularity. The surface also narrows in the orthogonal direction. 

Motor scaling shows a trend similar to the 1D analysis (\cref{fig:motor_scale_modeling}.) The surface resembles a belted ellipsoid which is characteristic of the effective mass ellipsoid from~\cite{khatib1995Inertial}. This belted ellipsoid elongates as we move away from direct-drive ($r_m = \SI{60}{mm}$). The inertia orthogonal to the second link is isolated by the spring at the second joint and  does one depend on the reflected inertia of the actuator. No change to the surface is seen in that direction.

At slower velocities, the collision reflex is dominated by the sensing phase where $I_{\Romannum{2}\to\Romannum{3}} \propto 1/v$, \cref{eq:sensing}. As the sensing phase does not depend on the inertia of the actuator, the surface approaches a circle -- isometric impulse in all directions. Increasing the velocity of impact has two effects: (1) the surface elongates perpendicular to the second link as initial impact (which depends on inertia, \cref{eq:initial}) starts to dominate; (2) the overall surface shrinks up to $v_0 = v^*$ and then expands (see \cref{eq:v_star}).

\section{Experiments}
\label{sec:experiments}
Through experiments we attempt to ground the collision reflex metric in reality and discover deviations from it. We implement the collision reflex experiment: systems capable of collision detection undergo an unexpected collision with a force-torque sensor (ATI mini40). The force signal from the sensor is captured and the collision reflex is calculated as the area under the force-time curve. \cref{fig:collision_reflex_plot} (bottom) shows a typical trace of a force-time curve from an experiment. The collision reflex is measured for: (\cref{sec:1dexp}) 1D collisions -- comparing a single joint geared and direct-drive actuator; and (\cref{sec:handexp}) Gripper collisions -- comparing the Schunk WSG-50 gripper and the DDHand.

\subsection{Collision Reflex in One-dimension}
\label{sec:1dexp}
\begin{figure}
  \centering
  \includegraphics[width=\columnwidth]{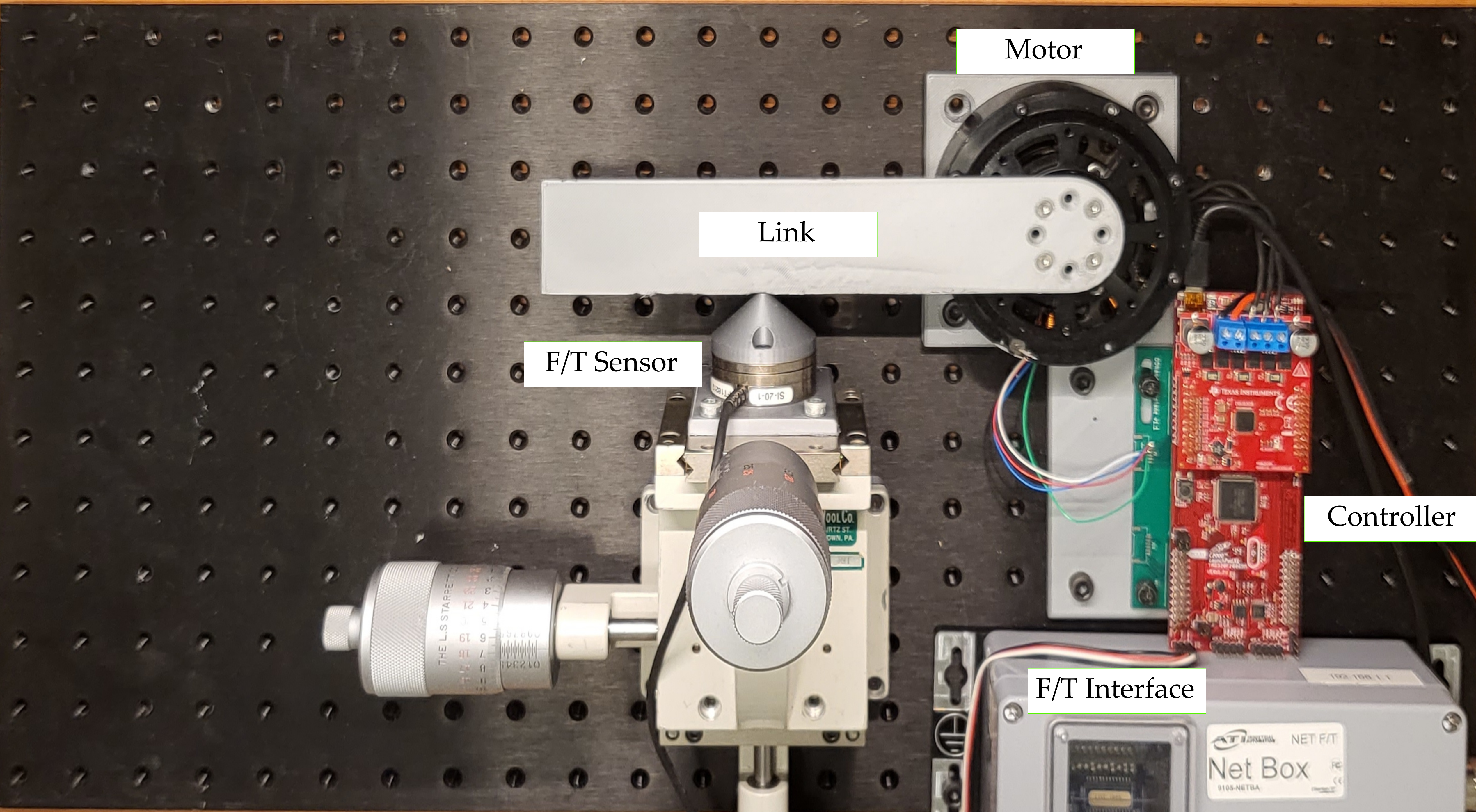}
  \caption{Experimental Setup for 1D experiments. The figure shows a gear motor (T-motor A8 with a transmission ratio of 9:1) connected to a link which impacts an ATI mini40 force-torque sensor. The motor is driven by a TI instaspin enabled C2000 microcontroller with a BOOSTXL-DRV8305 3 phase driver. The experiment is repeated with a T-motor U8 in a direct-drive configuration.}
  \label{fig:1d_setup}
\end{figure}

We test the collision reflex of a direct-drive and geared actuator in a one-dimensional collision. The direct-drive actuator is a T-motor U8 (a BLDC motor used, e.g., on the Ghost Robotics Minitaur\cite{kenneally2016Design}).  The geared actuator is the T-motor A8 which is a U8 motor augmented with a 9:1 planetary gearbox. Both motors are controlled by a TI instaspin motor controller running custom firmware with a RLS RM-08 magnetic encoder for position sensing and commutation. \cref{fig:1d_setup} shows the setup for this experiment. The collision is set up to occur at link angle of \SI{0}{rad} with a characteristic length of \SI{114.3}{mm}. 

\begin{figure*}
  \centering
  \includegraphics[width=0.8\textwidth]{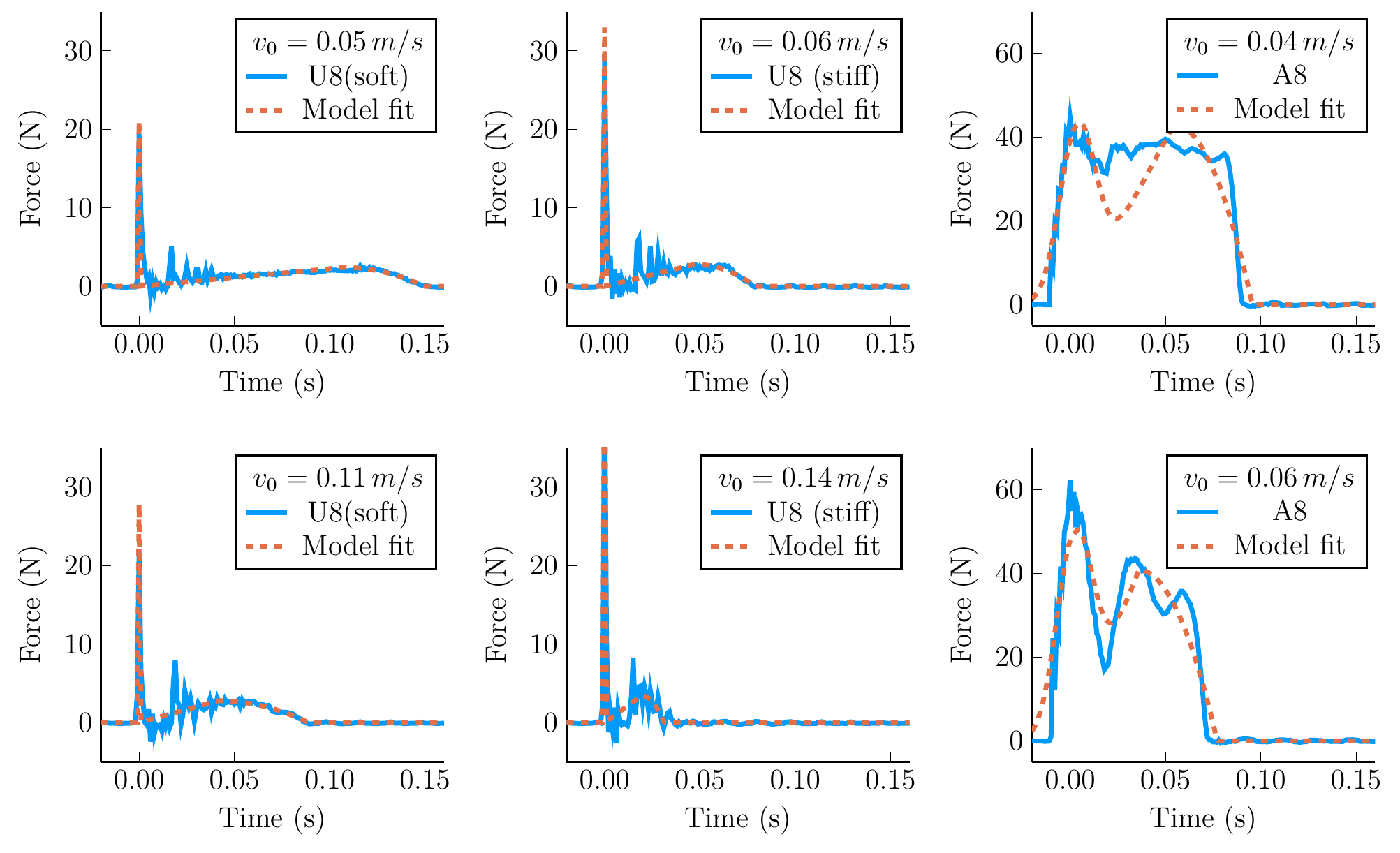}
  \caption{Force measurements over the duration of the impulse for: (left) U8 (direct-drive, soft), (center) U8 (direct-drive, stiff) and the A8 (geared) actuator configurations. The dotted line in each shows the collision reflex model fit to the data. ~\cref{tab:one_shot} shows the uncertainty in the parameter fit for each system.}
 
  \label{fig:1d_experiment}
\end{figure*}

\begin{figure}
  \centering
  \includegraphics[width=0.9\columnwidth]{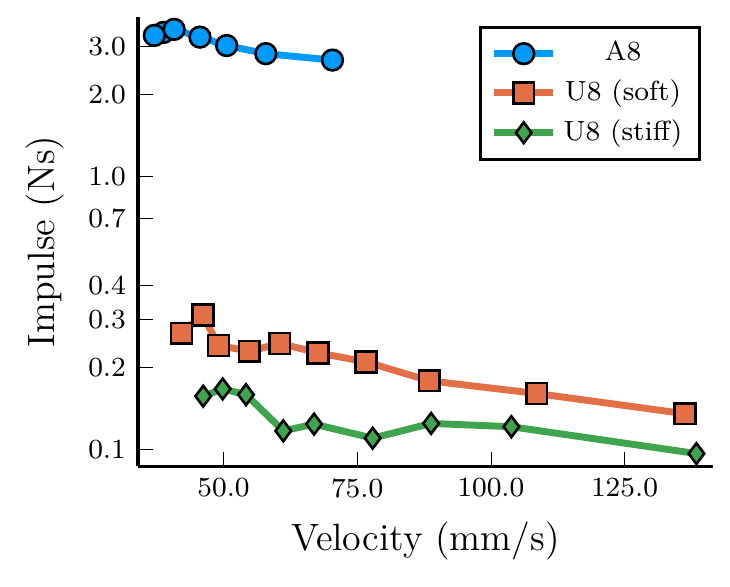}
  \caption{ The collision reflex impulse with varying velocity for the A8 (geared) and U8 (direct-drive). The U8 experiments are duplicated with two levels of software stiffness. The y-axis is shown as a log plot to better show the separation between the actuator configurations.}
  \label{fig:1d_experiment_nshot}
\end{figure}

The behavior for this experiment implements the sensing phase as a move-until-touch action. The link starts at initial pose $q_i = -\pi/2$ and executes a spline trajectory to final pose $q_f = \pi/2$ for time $t$. The velocity of impact is varied in proportion to the trajectory time. A threshold on force at the link, as measured by motor current, provides collision detection capabilities. Once collision is sensed, the motor is commanded to $q_i$ as fast as the system allows. For the short compression distance of the force sensor, the motor can be assumed to operate at constant acceleration.

The evolution of force over the impact is captured at \SI{3500}{Hz} over a UDP communication link. This evolution is shown in \cref{fig:collision_reflex_plot} (bottom) for a U8 motor and more examples are shown in~\cref{fig:1d_experiment} for the U8 in low and high gain stiffness, as well as the A8. Changing stiffness for the A8 has little effect on the output due to higher reflected inertias. We estimate the parameters of the collision reflex model ($m_f$, $k_s$, $F_s$ and $a$) for each collision using a one-shot least squares fit of the force-time profile. Mechanical Stiffness ($k_m$) is assumed to be \SI{2e7}{N/m} which is the mechanical stiffness of the F/T sensor along the sensing direction. The normal distributions of these parameters are reported in~\cref{tab:one_shot}. 

\begin{table}
  \centering
  \footnotesize
%  \maxsizebox{\columnwidth}{!}{
    \renewcommand\arraystretch{1.3}
    \setlength{\tabcolsep}{0.35em}
    \begin{tabular}{lcccc}
      \toprule
      & & \multicolumn{2}{c}{U8} &A8\\
      \cmidrule(){3-4}
      Parameter&Unit& Low $k$ (Soft) & High $k$ (Stiff) &\\
      \midrule
      Finger mass ($m_f$) &\si{kg}        &\num{0.1\pm 0.02}    &\num{0.145\pm 0.07}    &\num{20.04\pm 2.4}\\
      Stiffness ($k_s$)     &\si{N mm^{-1}}  &\num{0.440\pm 0.07}     &\num{1.07\pm 0.41}     &\num{23.7\pm 6.3}\\
      Threshold ($F_s$)   &\si{N}         &\num{3.00\pm 0.34}   &\num{3.03\pm 0.33}   &\num{40.85\pm 2.57}\\
      Acceleration ($a$)  &\si{mm s^{-2}}  &(\num{4.7\pm 3.1})E-2   &\num{7.0\pm 15} &\num{3.1\pm 1.2}\\

      \bottomrule
    \end{tabular}
 % }
  \caption{Estimated parameters from the 1-shot model fit for the Direct-drive (U8) and geared (A8) actuators. }
  \label{tab:one_shot}
\end{table}

The finger mass $m_f$ for the U8 in high and low stiffness configurations is $\approx$ \SI{0.12}{kg}  while that of the A8 is higher due to added inertia of the gearbox and the reflected inertia from the transmission. The stiffness increases as the U8 switches from low and high stiffness modes. The higher stiffness of the A8 is due to reduced backdrivability form the transmission. As the threshold $F_s$ is applied on the motor side of the transmission for proprioceptive sensing, the U8 actuator detects collision at $\approx$ \SI{3}{N} force while the A8 detects collision at $\approx$ \SI{40}{N}. \cref{fig:1d_experiment} shows the model fit to measure these parameters. Note that the model fit for the A8 motor is not as accurate due to deviations in the model from reality caused by additional inertia and mechanical delays due to the added transmission. 

The total impulse imparted in the collision is determined by numerical integration of force over time using adaptive Gauss-Kronrod quadrature~\cite{notaris2016gauss} as shown in~\cref{fig:1d_experiment_nshot}. We see that for both direct-drive (U8, $N=1$) and quasi-direct-drive (A8, $N=9$) actuator configurations the impulse decreases as we increase velocity in this regime. It is not surprising that the impulse for the A8 is much higher than that of the U8 due to higher reflected inertias of the U8 rotor through the transmission. Both systems are operating at $v_0 < v^*$ (\cref{eq:v_star}) in the above experiments.

\subsection{Comparing the Collision Reflex of Robot Hands}
\label{sec:handexp}
We measure the collision reflex of two robot hands: The Schunk WSG-50 gripper and the Dexterous DDHand  \cite{bhatia2019Direct}. 
The WSG-50 end-effector is a popular COTS offering from Schunk. It has a parallel jaw configuration driven by a electric servo motor with a ball screw transmission. It has a stroke of up to \SI{110}{mm} and a peak grip force of \SI{80}{N}. 
The Dexterous DDHand is a 4 degree-of-freedom direct-drive end-effector. It is actuated by a set of T-motor GB54-2 motors connected to 3 parallelogram linkages that mimic a two-link serial robot while keeping the fingertips parallel. 

\cref{fig:reflex_experiment} shows the setup for this experiment. The two grippers were mounted on a stationary ABB IRB 120 robot above a force-torque sensor (ATI Mini40.) A move-until touch controller was implemented on both robots with proprioceptive collision sensing. As the control architecture for the schunk gripper is abstracted away from the user, a workaround using the gripper's close until force command was used with one of the fingertips removed. Once the action reported successful completion, the robot was instructed to open at it's maximum speed.

For the DDHand, the collision sensing was implemented as an error threshold on the position controller, i.e. if the position tracking for any of the joints was lagging behind the command, the finger was declared in collision. Again, once the collision was detected, the robot was instructed to open at its maximum speed.

Data was collected at pre-impact velocities up to \SI{55}{mm/s} for the Schunk gripper and up to \SI{150}{mm/s} for the DDHand. The force-time data was integrated using adaptive Gauss-Kronrod quadrature and the results are presented in~\cref{fig:reflex_experiment}.

The DDHand, being direct-drive, is more transparent. It follows the unintuitive trend of lower impulse at higher pre-impact velocities (as all attempted collisions for the DDHand exist in the space of collisions where $v_0 < v^*$). The Schunk gripper on the other hand shows a more conventional linear trend due to the impulse from the initial impact phase dominating. All attempted collisions for the Schunk gripper exist in the sa  the space of collisions where $v_0 > v^*$.

\section{Discussion and Future Work}
\label{sec:discussion}

A robot's transparency and capability to mitigate collisions depends on structural parameters, actuators, sensing, and control architectures. This paper proposes the collision reflex metric to quantify this capability. The metric builds upon the feel-cage task and avoids the dependence on parameters of the external object to be caged. An analytical expression for the collision reflex is derived in one-dimension and extended to higher dimensions. 

The three phases of collision discussed in the paper play varied roles depending on the pre-impact velocity. At high velocities, the impulse during collision behaves as one would expect -- the initial impact dominates and the impulse intuitively follows the change in momentum of the reflected inertia at the contact point. The behavior at lower speeds is rather counter intuitive, as the sensing phase dominates the impulse. As you increase the pre-impact velocity, the total impulse decreases. We show that the operating regime where the impulse reduces as the speed increases, exists on real systems at relevant velocities. However, note that this behavior is highly dependent on the sensing and control architecture. 

Even so, no matter what the structure, sensing, and control are for a system, the metric is generalizable. The experiment to measure the collision reflex of any system, real or simulated, is simple. The system is impacted against a rigid surface and the area under the force-time curve for that event gives the collision reflex metric for that system.

The collision reflex metric considers the total impulse during a collision and not just the maximum force (as in~\cite{kirschner2021Experimental}). Why not use the maximum force then? The impulse metric gives you more information about the transparency as it depends both on the sensing bandwidth an actuation bandwidth. A metric based on maximum force will be easily defeated by mechanical compliance which has a negative effect on actuation bandwidth. 

The simple mass-spring-mass collision model used to analyze collision reflex has some limitations. For example, it doesn't account for the nonlinear behavior of transmissions that stem from backlash, friction, and gear inertias. The model also fails to capture time delays in communication, although those could be added in. In both our real-world experiments, the behavior of the geared systems deviated from the ideal response. It is unclear how much of this deviation can be accounted for with a first principles approach.
These shortcomings make it more difficult to use the model for learning parameters and accurate prediction of the collision reflex from a few data points. The model parameters learned from experiments (\cref{tab:one_shot}) have non-trivial uncertainties associated with them. Future work should address the limitations of the simple model in order to enable more accurate predictions.
% The general trend is nevertheless useful and may still be captured by the model. 

The inaccuracies in modeling the real world do not take away from the usefulness of the collision reflex metric. The analysis in this paper has shown some unintuitive results for closed-loop robotic systems. Most notably, there is an optimal velocity for minimizing impulses during collision. The metric also captures the trends that we see in real-world impulses during collision. The use of this metric is relevant in the design and motion planning of robots where minimization of impulses is essential in a collision.

\bibliographystyle{IEEEtran}
\bibliography{references, other_references}

\end{document}